\pdfoutput=1
\documentclass{article}

\usepackage{iclr26/iclr2026_conference,times}

\usepackage{amsmath,amsfonts,bm}

\def\eqref#1{equation~\ref{#1}}

\def\1{\bm{1}}

\DeclareMathAlphabet{\mathsfit}{\encodingdefault}{\sfdefault}{m}{sl}
\SetMathAlphabet{\mathsfit}{bold}{\encodingdefault}{\sfdefault}{bx}{n}

\usepackage{hyperref}
\hypersetup{
  pdftitle={LitTraceQA: A Benchmark for Multi-Stage Grounding and Verification in Scientific Question Answering},
  pdfauthor={Xuye Liu, Yimu Wang, Peng Shi, Bo Xue, Xiangrui Ke, Songcheng Cai, Kath Choi, Di Wu, Freda Shi, Krzysztof Czarnecki}
}
\usepackage{url}
\usepackage{booktabs}
\usepackage{xcolor}
\usepackage{graphicx}
\usepackage{amsmath}
\usepackage{amssymb}
\usepackage{enumitem}
\usepackage{pifont}
\usepackage{tikz}
\usetikzlibrary{arrows.meta,positioning,fit,shapes.geometric}

\iclrfinalcopy

\definecolor{mpqaBlue}{RGB}{41,98,160}
\definecolor{mpqaGreen}{RGB}{69,142,94}
\definecolor{mpqaOrange}{RGB}{214,126,46}
\definecolor{mpqaGray}{RGB}{92,100,112}
\definecolor{litqaCheck}{RGB}{0,166,81}
\definecolor{litqaPartial}{RGB}{245,140,35}
\definecolor{litqaCross}{RGB}{218,45,45}
\newcommand{\cmark}{\textcolor{litqaCheck}{\ding{51}}}
\newcommand{\pmark}{\textcolor{litqaPartial}{\ensuremath{\blacktriangle}}}
\newcommand{\xmark}{\textcolor{litqaCross}{\ding{55}}}

\newcommand{\MPQASamples}{4,978}
\newcommand{\MPQAUniquePapers}{4,859}
\newcommand{\MPQAGoldLinks}{8,612}
\newcommand{\MPQAMultiPaper}{3,228}
\newcommand{\MPQAMultiPaperPct}{64.85\%}
\newcommand{\MPQASinglePaper}{1,750}
\newcommand{\MPQASinglePaperPct}{35.15\%}
\newcommand{\MPQAMeanPapers}{1.73}
\newcommand{\MPQAMeanEvidence}{2.62}

\newcommand{\statbar}[4]{%
  #1 & #2 & \begin{tikzpicture}[baseline=-0.5ex]
    \fill[mpqaBlue!18] (0,0) rectangle (3.8,0.17);
    \fill[mpqaBlue!75] (0,0) rectangle (#3,0.17);
  \end{tikzpicture} \\
}

\newcommand{\MPQAVenueBars}{%
\statbar{ICCV 2025}{4,187 (48.6\%)}{3.80}{4,187}
\statbar{NAACL 2025}{972 (11.3\%)}{0.88}{972}
\statbar{ACL 2025}{925 (10.7\%)}{0.84}{925}
\statbar{ICLR 2025}{443 (5.1\%)}{0.40}{443}
\statbar{ICML 2025}{438 (5.1\%)}{0.40}{438}
\statbar{CVPR 2025}{436 (5.1\%)}{0.40}{436}
\statbar{ECCV 2024}{408 (4.7\%)}{0.37}{408}
\statbar{NeurIPS 2025}{404 (4.7\%)}{0.37}{404}
\statbar{EMNLP 2025}{399 (4.6\%)}{0.36}{399}
}

\title{\raggedright LitTraceQA: A Benchmark for Multi-Stage Grounding
and Verification in Scientific Question Answering}

\author{\begin{minipage}[t]{\dimexpr\textwidth-2\tabcolsep\relax}
\raggedright\bfseries
Xuye Liu\textsuperscript{1} \quad Yimu Wang\textsuperscript{1} \quad Peng Shi\textsuperscript{1} \quad
Bo Xue\textsuperscript{2} \quad Xiangrui Ke\textsuperscript{1} \\
Songcheng Cai\textsuperscript{1} \quad Kath Choi\textsuperscript{1} \quad Di Wu\textsuperscript{3} \quad
Freda Shi\textsuperscript{1} \quad Krzysztof Czarnecki\textsuperscript{1} \\[0.45em]
\normalfont\small
\textsuperscript{1}University of Waterloo \quad
\textsuperscript{2}City University of Hong Kong \quad
\textsuperscript{3}University of Amsterdam
\end{minipage}}

\begin{document}

\maketitle
\lhead{}

\begin{abstract}
Scientific literature is increasingly used as a knowledge source for language
models, retrieval-augmented generation systems, and research assistants, but
answering research questions from papers requires more than fluent generation.
A reliable system must identify the relevant papers, locate the concrete
evidence that supports the answer, and produce a response that is faithful to
that evidence. We present LitTraceQA, a benchmark for literature-grounded
question answering over scientific papers. Given a research question and a
metadata pool of papers, a system must return three connected outputs:
canonical paper identifiers, supporting evidence locations, and answers in one
or more requested formats, including free-form text, multiple-choice answers,
and structured tables. LitTraceQA targets evidence types common in scientific
reading: tables, figures, text spans, equations or algorithms, and citation
contexts. The public development split contains 55 examples, including 26
hidden-source single-paper questions and 29 multi-paper questions, and provides
gold papers, evidence annotations, and answers for local validation. We also
analyze a larger final annotation collection with \MPQASamples{}
unique-question records over \MPQAUniquePapers{} unique gold papers.
By evaluating paper retrieval, evidence grounding, and answer accuracy
separately, LitTraceQA provides a testbed for scientific QA systems that produce
verifiable answers rather than unsupported summaries.
\end{abstract}

\section{Introduction}

Scientific literature is increasingly used as an external knowledge source for language models, retrieval-augmented generation systems, and research assistants \citep{lewis2020retrieval}. However, answering research questions from papers is not simply a matter of generating fluent summaries: systems must recover information distributed across full papers and provide concrete evidence for their answers \citep{dasigi2021qasper}. A reliable system must first identify the relevant papers, then locate the specific evidence that supports the answer, and finally produce a response that is faithful to that evidence \citep{gao2023alce}. This process is especially important in scientific domains, where answers often depend on exact benchmark numbers, experimental settings, figure trends, algorithmic details, or citation contexts \citep{dasigi2021qasper,pramanick2024spiqa}.

Existing question answering benchmarks often focus on answer correctness alone,
or they provide the source document in advance. Recent scientific QA resources
make important progress on individual pieces of the problem: M3SciQA evaluates
multimodal, multi-document scientific QA over paper clusters~\citep{li2024m3sciqa};
LitSearch focuses on realistic scientific literature retrieval queries~\citep{ajith2024litsearch};
AirQA evaluates academic-paper QA with multiple question types and
instance-level evaluation functions~\citep{huang2025airqa}; and ResearchQA
scales scholarly QA with survey-mined queries and rubrics~\citep{yifei2025researchqa}.
Grounded visual benchmarks such as VISTAQA and view-level evidence
identification show that answer correctness can hide evidence-selection
failures when grounding is not measured explicitly~\citep{azadani2026vistaqa,
wang2026viewevidence}. These settings leave open a central challenge for
literature-grounded reasoning: can a model recover the right papers from a
large paper pool, identify the evidence inside those papers, and generate an
answer whose claims can be checked?

We introduce LitTraceQA, a benchmark and shared-task development set for
literature-grounded question answering. Each example is organized around a
research question. The system receives the question and a paper metadata pool,
and must return three outputs: the relevant paper identifiers, supporting
evidence locations, and the final answer. LitTraceQA is designed to evaluate the
complete chain from retrieval to evidence grounding to answer generation, rather
than treating citations or evidence as optional explanations after the answer
has already been produced.

LitTraceQA emphasizes evidence types that appear frequently in scientific
papers and that can be evaluated in a reproducible way. The benchmark focuses
on five primary evidence types: tables, figures, text spans, equations or
algorithms, and citation contexts. These categories cover common scientific
reading behaviors, such as comparing reported numbers across papers, reading
trends from figures, extracting method or setting descriptions from prose,
interpreting formal procedures, and understanding how papers position
themselves relative to prior work.

\begin{figure}[t]
\centering
\includegraphics[width=\linewidth]{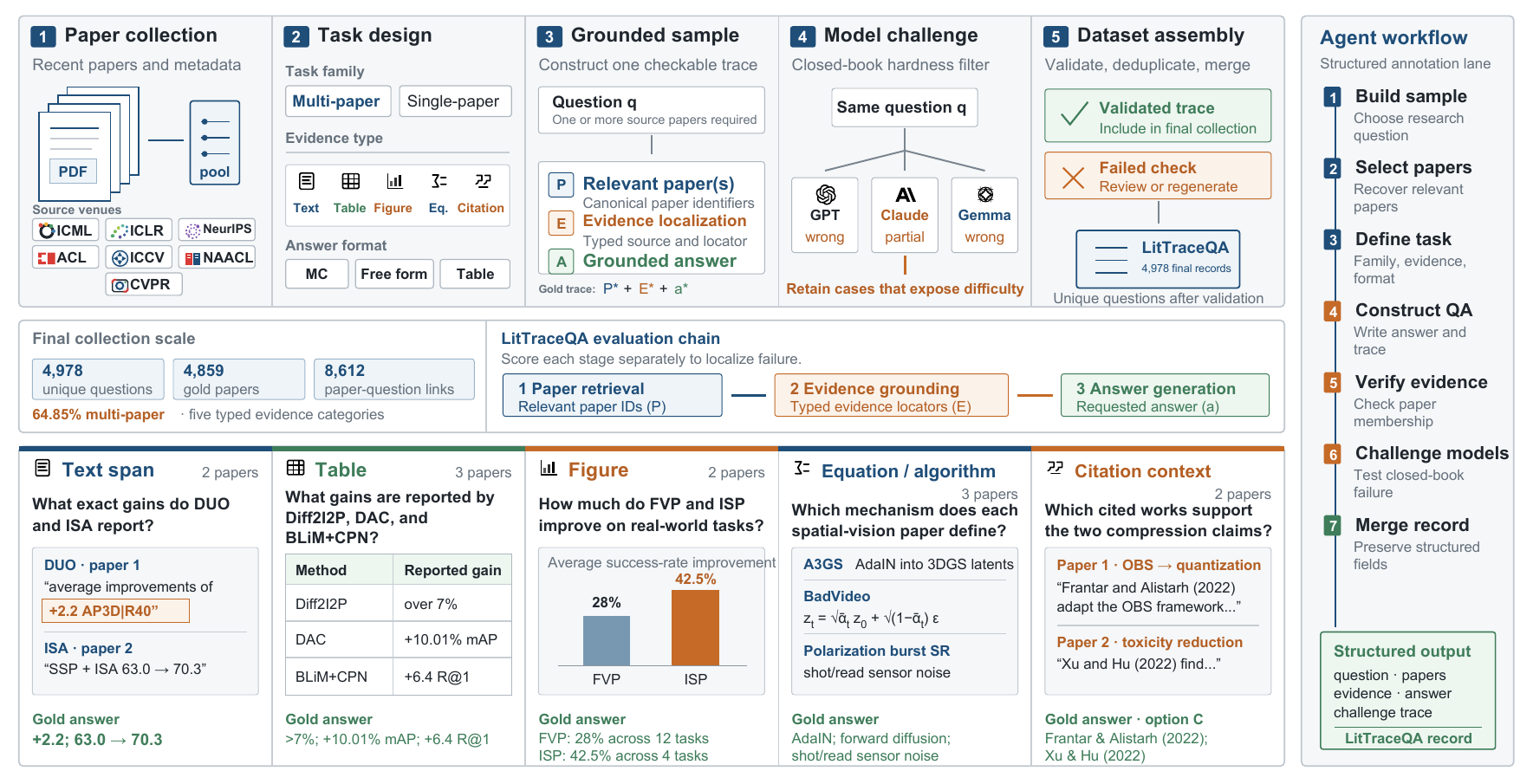}
\caption{LitTraceQA constructs literature-grounded QA examples as traceable chains from paper retrieval to typed evidence grounding and final answer generation. The top row summarizes collection, task design, grounded construction, closed-book model challenge, and validation; compact vector marks identify source venues, evidence types, and the GPT, Claude, and Gemma challengers. The middle strip states the three-stage evaluation contract, the bottom row gives examples from five scientific evidence types, and the right lane summarizes the structured annotation workflow.}
\label{fig:task-example}
\end{figure}

Figure~\ref{fig:task-example} summarizes the benchmark scope. It follows the
paper-overview style used in recent academic QA benchmarks: the left side maps
the task space and evidence categories, while the right side shows the trace
bottleneck that answer-only evaluation misses. LitTraceQA requires the
response trace to be correct at all three levels. A model can fail because it
misses the paper, because it finds the paper but not the evidence, or because it
finds the evidence but still produces the wrong answer.

The public development split contains 55 examples, including 26 hidden-source
single-paper questions and 29 multi-paper questions. It covers all five primary
evidence types, with 17 table-grounded questions, 15 figure-grounded questions,
9 text-span questions, 7 equation-or-algorithm questions, and 7
citation-context questions. Although the public split is compact, it provides
the complete task format for participants to build data loaders, tune prompts
or systems, and validate submissions before evaluation on a hidden test set.

This paper makes three contributions. First, we define a literature-grounded QA
task that explicitly separates paper retrieval, evidence grounding, and answer
generation. Second, we provide annotations over diverse scientific evidence
types, including non-textual and semi-structured sources such as tables and
figures. Third, we support multiple answer formats, allowing the same framework
to evaluate free-form synthesis, multiple-choice decisions, and structured
comparison tables. We further report an audit of a larger local annotation
collection to characterize the distributional profile and release requirements
for scaling LitTraceQA.

\section{Related Work}

\paragraph{Question answering over scientific papers.}
Early scientific-document QA benchmarks already recognized that research papers
are not ordinary short documents. QASPER asks information-seeking questions over
NLP papers and includes supporting evidence for answers~\citep{dasigi2021qasper}.
QASA studies more advanced questions over AI and machine-learning articles and
emphasizes evidence-based reasoning over full papers~\citep{lee2023qasa}.
Recent multimodal and artifact-centered datasets extend this line: M3SciQA
targets multimodal, multi-document scientific QA over paper
clusters~\citep{li2024m3sciqa}; SPIQA focuses on figures and tables in
scientific papers~\citep{pramanick2024spiqa}; and SCITAT studies scientific
tables and text with diverse reasoning types~\citep{zhang2025scitat}. AirQA
further evaluates academic-paper QA with single-document, multi-document,
retrieval, and comprehensive QA settings~\citep{huang2025airqa}. LitTraceQA
is aligned with these efforts in treating scientific papers as structured,
multi-artifact documents, but it makes the answer trace explicit at the level of
paper identifiers and typed evidence items.

\paragraph{Scientific literature retrieval and synthesis.}
A second line of work evaluates whether systems can find, rank, and synthesize
scientific literature. LitSearch provides realistic literature-search queries
for scientific retrieval~\citep{ajith2024litsearch}. Scientific representation
and corpus resources such as S2ORC and SciRepEval support broader text mining,
ranking, and search tasks over scholarly documents~\citep{lo2020s2orc,
singh2022scirepeval}. Retrieval-augmented systems such as PaperQA and
OpenScholar show how scientific QA can be framed as an evidence-backed
literature-synthesis problem over large paper collections~\citep{lala2023paperqa,
asai2024openscholar}. ResearchQA similarly evaluates scholarly long-form QA at
scale using survey-derived queries and rubrics~\citep{yifei2025researchqa}.
LitTraceQA differs in its output contract: instead of asking for only a
long-form synthesis or a ranked paper list, it asks for the relevant papers,
the typed evidence objects, and a checkable answer.

\paragraph{Evidence, attribution, and scientific verification.}
Scientific claim verification and evidence extraction provide a closely related
view of grounding. SciFact asks systems to retrieve scientific abstracts and
identify rationales that support or refute expert-written claims~\citep{wadden2020scifact}.
Evidence Inference studies treatment-comparison prompts over clinical-trial
articles and requires supporting evidence for inferred results~\citep{deyoung2020evidenceinference}.
More general attribution benchmarks ask whether generated answers are properly
supported by cited sources. Attributed QA formalizes answer attribution as an
evaluation problem for large language models~\citep{bohnet2022attributedqa},
while ALCE evaluates citation-supported generation along fluency, correctness,
and citation-quality dimensions~\citep{gao2023alce}. In visual domains,
VISTAQA evaluates joint visual QA and pixel-level evidence~\citep{azadani2026vistaqa},
and view-level evidence identification separates view selection from answer
generation in multi-view settings~\citep{wang2026viewevidence}. LitTraceQA
brings this attribution perspective into scientific-paper QA: the evidence unit
is not a generic passage alone, but a typed paper artifact such as a table,
figure, text span, equation or algorithm, or citation context.

\begin{table}[t]
\centering
\scriptsize
\caption{Component-level comparison with related scientific QA datasets and retrieval benchmarks. A checkmark indicates that the component is an explicit output or evaluation target; a triangle indicates partial or implicit support. LitTraceQA is the only benchmark in this comparison that combines paper retrieval, multi-paper QA, typed artifact evidence, checkable answers, strict trace evaluation, and closed-book hardness filtering.}
\label{tab:related}
\setlength{\tabcolsep}{3.0pt}
\begin{tabular}{p{0.16\linewidth}ccccccp{0.18\linewidth}}
\toprule
Benchmark & Paper ret. & Multi paper & Typed evid. & Checkable ans. & Strict trace & Hardness & Primary limitation relative to LitTraceQA \\
\midrule
QASPER & \xmark & \xmark & \pmark & \xmark & \pmark & \xmark & Given-paper QA; evidence is not a retrieval output. \\
QASA & \xmark & \xmark & \pmark & \xmark & \pmark & \xmark & Full-paper reasoning without paper-pool retrieval. \\
M3SciQA & \pmark & \cmark & \cmark & \xmark & \pmark & \xmark & Multimodal QA, but not a strict paper--evidence--answer trace. \\
SPIQA & \xmark & \xmark & \cmark & \xmark & \pmark & \xmark & Figure/table QA with the paper context largely given. \\
SCITAT & \xmark & \xmark & \cmark & \cmark & \pmark & \xmark & Table-text reasoning, not retrieval-grounded paper QA. \\
AirQA & \cmark & \cmark & \cmark & \pmark & \pmark & \xmark & Broad QA settings, but no closed-book hardness-filtered MC/table trace. \\
LitSearch & \cmark & \xmark & \xmark & \xmark & \xmark & \xmark & Retrieval benchmark without evidence or answer generation. \\
ResearchQA & \pmark & \cmark & \pmark & \xmark & \xmark & \xmark & Long-form scholarly QA with rubric-based assessment. \\
\textbf{LitTraceQA} & \cmark & \cmark & \cmark & \cmark & \cmark & \cmark & Paper IDs, typed evidence, and answer are jointly required. \\
\bottomrule
\end{tabular}
\end{table}

\section{Task Definition and Design Goals}

LitTraceQA is a traceable literature-grounded QA task over a fixed paper
metadata pool. The input contains a stable
`query\_id`, the natural-language `question`, and requested `answer\_types`;
multiple-choice inputs additionally expose the option set, and table-answer
inputs expose the table schema used for row alignment. The system must return
three connected outputs: canonical paper identifiers from the pool, evidence
locations inside those papers, and the requested answer components. Let
$\mathcal{P}$ be the released paper pool. Given a question $q$, a system returns
a set of relevant paper identifiers $\hat{P}$, a set of typed evidence items
$\hat{E}$, and an answer $\hat{a}$:
\begin{align}
    \hat{P} &= \{\hat{p}_1,\ldots,\hat{p}_m\},\\
    \hat{E} &= \{\hat{e}_1,\ldots,\hat{e}_n\},\\
    \hat{a} &= \mathrm{Answer}(q,\hat{P},\hat{E}).
\end{align}
The gold record contains `gold\_papers`, `evidence`, and `answer`. A gold paper
stores `paper\_id`, title, venue, and year. Each evidence item stores an
`evidence\_id`, the supporting `paper\_id`, a `source\_type`, the evidence text
or value, and a locator. The public scorer uses coarse locators: page plus table
ID for tables, page plus figure ID for figures, and page-level locations for
text spans, equations or algorithms, and citation contexts. Richer fields such
as row, column, section, region, equation ID, algorithm step, or cited paper can
appear as annotation detail, but the task does not require systems to reproduce
every fine-grained field.

The task is organized by three orthogonal schema axes. First, `task\_family`
specifies whether the question is
`hidden\_source\_single\_paper`, where one relevant paper must be recovered from
the pool, or `multi\_paper`, where the answer depends on evidence from multiple
papers. Second, `primary\_evidence\_type` identifies the main source type:
table, figure, text span, equation or algorithm, citation context, and, in the
schema, paper summary. Third, `answer\_types` specifies whether the answer is
free-form, multiple choice, a structured table, or a combination of these
components. The public LitTraceQA development split uses this full schema and
contains free-form, multiple-choice, and structured-table answer components. The
larger local LitTraceQA annotation collection analyzed in this paper uses the
same core trace fields and currently restricts final answers to multiple-choice
and structured-table formats to make automatic checking more reproducible
during data scaling.

\begin{table}[t]
\centering
\scriptsize
\caption{Representative hard LitTraceQA annotation examples selected from the local collection. Each row is a coherent multi-paper question whose answer requires evidence from more than one gold paper, rather than unrelated single-paper subquestions. Complete annotation cards are provided in Appendix~\ref{app:annotation-examples}.}
\label{tab:task-examples}
\setlength{\tabcolsep}{2.0pt}
\renewcommand{\arraystretch}{1.15}
\begin{tabular}{p{0.13\linewidth}p{0.27\linewidth}p{0.25\linewidth}p{0.27\linewidth}}
\toprule
Primary evidence / answer & Coherent question sample & Gold evidence used & Gold answer and observed failure \\
\midrule
Figure + table &
Two robot manipulation policy papers: compare the real-world success-rate improvement and number of real-world tasks for FVP and Image-to-Sphere Policy. &
FVP uses Figure~1 context and abstract text; Image-to-Sphere uses its introduction and Figure~1 caption. &
FVP: 28\% over 12 real-world tasks; Image-to-Sphere Policy: 42.5\% over 4 real-world tasks. All three challenger models filled only 1/5 checked cells correctly. \\
\addlinespace
Table-style metric comparison + table &
Three retrieval papers: align the headline gain reported by Diff2I2P, DAC, and BLiM+CPN across registration, open-set 3D retrieval, and text-video retrieval. &
Each answer row is grounded in the corresponding paper's abstract-level result sentence, with an additional DAC context span. &
Diff2I2P: over 7\% registration-recall gain on 7-Scenes; DAC: +10.01\% mAP over four open-set 3DOR datasets; BLiM+CPN: 6.4 R@1 average gain. All challenger models returned much smaller or mismatched gains. \\
\addlinespace
Text span + table &
Two inference-time adaptation papers: extract the headline quantitative gain for DUO on KITTI and ISA on the SSP baseline. &
DUO requires the introduction sentence giving +2.2 AP3D$\vert$R40 on KITTI; ISA requires the overview value ``SSP + ISA 63.0 $\rightarrow$ 70.3''. &
DUO: +2.2 AP3D$\vert$R40 in the Car category; ISA on SSP: 63.0 $\rightarrow$ 70.3. All challenger models produced plausible but wrong gains such as +5.75 or +7.9. \\
\addlinespace
Eq./alg. + table &
Three spatial-vision papers: compare the exact noise-feature injection or noise-modeling mechanism defined by A3GS, BadVideo, and polarization burst SR. &
A3GS supplies the AdaIN injection operator; BadVideo supplies the forward diffusion noising equation; polarization burst SR supplies the rejected Gaussian-noise assumption and replacement sensor model. &
AdaIN for 3DGS latent style injection; $z_t=\sqrt{\bar{\alpha}_t}z_0+\sqrt{1-\bar{\alpha}_t}\epsilon$, $\epsilon\sim\mathcal{N}(0,1)$; reject additive white Gaussian noise in Stokes components in favor of shot/read sensor noise. All challenger models missed the row-level match. \\
\addlinespace
Citation context + MC &
Two LLM-compression papers: identify which cited works are used for OBS-to-quantization adaptation and for toxicity reduction under knowledge distillation. &
One related-work citation context comes from self-calibration quantization; the other comes from the inference-acceleration bias paper. &
Gold option C: Frantar and Alistarh (2022) for OBS-to-quantization; Xu and Hu (2022) for monotonic toxicity reduction in GPT-2. All challenger models selected option A, confusing later GPTQ/SparseGPT-style citations with the cited source. \\
\bottomrule
\end{tabular}
\renewcommand{\arraystretch}{1.0}
\end{table}

\paragraph{Hidden-source single-paper QA.}
In this family, the question does not give the answer-bearing paper ID. The
system must recover the paper from the metadata pool and then ground the answer
inside that paper. This setting is close to full-paper QA benchmarks such as
QASPER and QASA, but it adds a retrieval step and requires the output paper ID
to be checkable~\citep{dasigi2021qasper,lee2023qasa}. A question may ask for a
number in a table, a figure reading, an experimental setting, or a textual
claim; the answer is only valid if the retrieved paper and evidence locator also
match.

\paragraph{Multi-paper QA.}
In this family, the gold set contains multiple papers. The model must align
evidence across papers rather than solve independent single-paper questions.
Typical examples compare methods under a shared benchmark, reconcile values
reported in different tables or figures, or identify a prior work cited by more
than one paper. This family is motivated by multi-document scientific QA
resources such as M3SciQA~\citep{li2024m3sciqa}, but LitTraceQA makes the
paper set, evidence items, and final answer jointly visible to the evaluator.

\paragraph{Typed evidence grounding.}
Scientific-paper QA fails in different ways depending on the evidence source. A
table question may require a row and column value; a figure question may require
reading a plotted bar or caption; a text-span question may depend on an
implementation detail; an equation-or-algorithm question may depend on a
hyperparameter, update rule, or algorithm step; and a citation-context question
may depend on how a paper names or uses prior work. LitTraceQA preserves this
source type in `primary\_evidence\_type` and in every evidence item so that
errors in grounding can be diagnosed separately from errors in answer wording.

\paragraph{Answer formats.}
The schema supports free-form, multiple-choice, and table answers. Multiple
choice is useful when the evidence determines one discrete option, and the
submitted answer is the option letter. Structured tables are used when the
answer aligns multiple papers, methods, datasets, metrics, or settings; row-key
columns define how predictions are matched to gold rows. Free-form answers
appear in the public LitTraceQA development split, while the local LitTraceQA
collection analyzed below uses multiple choice and table answers only. This
difference is intentional: the release schema remains general, while the local
collection prioritizes automatically checkable answer formats.

The benchmark is guided by four design goals. \textbf{G1: schema fidelity.} The
paper describes the same core record contract used by the public LitTraceQA
release and the local LitTraceQA annotations: question, answer types, gold
papers, typed evidence, and answer. \textbf{G2: trace diagnostics.} Evaluation
should separate paper retrieval, evidence grounding, and answer correctness so
that a fluent but unsupported answer is not counted as fully correct.
\textbf{G3: scientific evidence coverage.} The data should cover tables,
figures, text spans, equations or algorithms, and citation contexts rather than
reduce paper QA to paragraph extraction. \textbf{G4: scalable quality control.}
The construction process should keep the trace fields machine-checkable,
validate evidence membership, and, for the local collection, filter for
questions that closed-book challenger models cannot answer reliably.

\section{Dataset Construction}

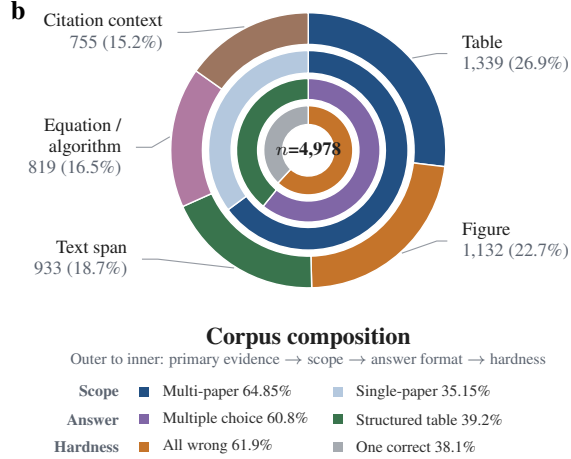
\begin{figure*}[t]
\centering
\begingroup
\colorlet{litBlue}{mpqaBlue!82!black}
\colorlet{litBlueLight}{mpqaBlue!35}
\colorlet{litGreen}{mpqaGreen!82!black}
\colorlet{litOrange}{mpqaOrange!92!black}
\colorlet{litGray}{mpqaGray!55}
\definecolor{litPurple}{RGB}{128,102,166}
\definecolor{litMagenta}{RGB}{176,122,161}
\definecolor{litBrown}{RGB}{156,117,95}

\begin{minipage}[t]{0.425\linewidth}
\vspace{0pt}
\textbf{a}\hspace{0.45em}\textbf{Final collection statistics}\par
\vspace{0.35em}
\scriptsize
\setlength{\tabcolsep}{1.8pt}
\renewcommand{\arraystretch}{1.00}
\begin{tabular*}{\linewidth}{@{\extracolsep{\fill}}lrr@{}}
\toprule
\textbf{Statistic} & \textbf{Count} & \textbf{Share} \\
\midrule
\multicolumn{3}{@{}l}{\textcolor{mpqaGray}{\bfseries Collection scale}} \\
Unique questions & \MPQASamples{} & -- \\
Unique gold papers & \MPQAUniquePapers{} & -- \\
Gold-paper links & \MPQAGoldLinks{} & -- \\
Mean papers / question & \MPQAMeanPapers{} & -- \\
Mean evidence / question & \MPQAMeanEvidence{} & -- \\
\addlinespace[0.15em]
\multicolumn{3}{@{}l}{\textcolor{mpqaGray}{\bfseries Question scope}} \\
Multi-paper questions & \MPQAMultiPaper{} & \MPQAMultiPaperPct{} \\
Single-paper questions & \MPQASinglePaper{} & \MPQASinglePaperPct{} \\
\addlinespace[0.15em]
\multicolumn{3}{@{}l}{\textcolor{mpqaGray}{\bfseries Answer format}} \\
Multiple choice & 3,029 & 60.8\% \\
Structured table & 1,949 & 39.2\% \\
\addlinespace[0.15em]
\multicolumn{3}{@{}l}{\textcolor{mpqaGray}{\bfseries Primary evidence}} \\
Table & 1,339 & 26.9\% \\
Figure & 1,132 & 22.7\% \\
Text span & 933 & 18.7\% \\
Equation / algorithm & 819 & 16.5\% \\
Citation context & 755 & 15.2\% \\
\addlinespace[0.15em]
\multicolumn{3}{@{}l}{\textcolor{mpqaGray}{\bfseries Closed-book hardness}} \\
All challengers wrong & 3,080 & 61.9\% \\
One challenger correct & 1,898 & 38.1\% \\
\bottomrule
\end{tabular*}
\end{minipage}
\hfill
\begin{minipage}[t]{0.545\linewidth}
\vspace{0pt}
\textbf{b}\par
\vspace{-0.65em}
\centering
\begin{tikzpicture}[
    x=1cm,
    y=1cm,
    callout/.style={font=\scriptsize, anchor=west, align=left, text=black!88},
    calloutleft/.style={font=\scriptsize, anchor=east, align=right, text=black!88},
    ringkey/.style={font=\tiny, anchor=west, text=black!82},
    ringname/.style={font=\tiny\bfseries, anchor=east, text=mpqaGray},
    centerlabel/.style={font=\scriptsize\bfseries, align=center, text=black!88},
    leader/.style={draw=mpqaGray!65, line width=0.42pt}
]
\newcommand{\ringslice}[7]{%
  \path[fill=#5, draw=white, line width=0.55pt]
    ([shift={(#1,#2)}]#3:#6)
    arc[start angle=#3,end angle=#4,radius=#6]
    -- ([shift={(#1,#2)}]#4:#7)
    arc[start angle=#4,end angle=#3,radius=#7]
    -- cycle;
}
\newcommand{\keyentry}[4]{%
  \fill[#3] (#1,#2) rectangle ++(0.12,0.12);
  \node[ringkey] at (#1+0.18,#2+0.06) {#4};
}

\ringslice{3.55}{4.02}{90}{-6.84}{litBlue}{1.82}{1.40}
\ringslice{3.55}{4.02}{-6.84}{-88.56}{litOrange}{1.82}{1.40}
\ringslice{3.55}{4.02}{-88.56}{-155.88}{litGreen}{1.82}{1.40}
\ringslice{3.55}{4.02}{-155.88}{-215.28}{litMagenta}{1.82}{1.40}
\ringslice{3.55}{4.02}{-215.28}{-270}{litBrown}{1.82}{1.40}

\ringslice{3.55}{4.02}{90}{-143.46}{litBlue}{1.32}{1.02}
\ringslice{3.55}{4.02}{-143.46}{-270}{litBlueLight}{1.32}{1.02}

\ringslice{3.55}{4.02}{90}{-128.88}{litPurple}{0.95}{0.67}
\ringslice{3.55}{4.02}{-128.88}{-270}{litGreen}{0.95}{0.67}

\ringslice{3.55}{4.02}{90}{-132.84}{litOrange}{0.59}{0.34}
\ringslice{3.55}{4.02}{-132.84}{-270}{litGray}{0.59}{0.34}
\node[centerlabel] at (3.55,4.02) {$n$=4,978};

\node[callout] at (5.46,5.30) {Table\\\textcolor{mpqaGray}{1,339 (26.9\%)}};
\draw[leader] ([shift={(3.55,4.02)}]41.58:1.84) -- (5.25,5.30) -- (5.39,5.30);

\node[callout] at (5.46,2.82) {Figure\\\textcolor{mpqaGray}{1,132 (22.7\%)}};
\draw[leader] ([shift={(3.55,4.02)}]-47.70:1.84) -- (5.25,2.82) -- (5.39,2.82);

\node[calloutleft] at (1.28,2.58) {Text span\\\textcolor{mpqaGray}{933 (18.7\%)}};
\draw[leader] ([shift={(3.55,4.02)}]-122.22:1.84) -- (1.56,2.58) -- (1.35,2.58);

\node[calloutleft] at (1.20,4.06) {Equation /\\algorithm\\\textcolor{mpqaGray}{819 (16.5\%)}};
\draw[leader] ([shift={(3.55,4.02)}]-185.58:1.84) -- (1.48,4.06) -- (1.27,4.06);

\node[calloutleft] at (1.75,5.58) {Citation context\\\textcolor{mpqaGray}{755 (15.2\%)}};
\draw[leader] ([shift={(3.55,4.02)}]-242.64:1.84) -- (1.96,5.58) -- (1.82,5.58);

\node[font=\footnotesize\bfseries, text=black!88] at (3.55,1.54) {Corpus composition};
\node[font=\tiny, text=mpqaGray] at (3.55,1.24)
{Outer to inner: primary evidence $\rightarrow$ scope $\rightarrow$ answer format $\rightarrow$ hardness};

\node[ringname] at (1.17,0.82) {Scope};
\keyentry{1.32}{0.76}{litBlue}{Multi-paper 64.85\%}
\keyentry{3.88}{0.76}{litBlueLight}{Single-paper 35.15\%}

\node[ringname] at (1.17,0.46) {Answer};
\keyentry{1.32}{0.40}{litPurple}{Multiple choice 60.8\%}
\keyentry{3.88}{0.40}{litGreen}{Structured table 39.2\%}

\node[ringname] at (1.17,0.10) {Hardness};
\keyentry{1.32}{0.04}{litOrange}{All wrong 61.9\%}
\keyentry{3.88}{0.04}{litGray}{One correct 38.1\%}
\end{tikzpicture}
\end{minipage}
\endgroup
\caption{LitTraceQA corpus statistics and composition. The left table reports exact counts and shares for the 4,978 unique-question records; the right concentric chart integrates primary evidence type in the outer ring with question scope, answer format, and closed-book hardness in successive inner rings. Multi-paper questions constitute 64.85\% of the collection, and all closed-book challengers fail on 61.9\% of records. Closed-book outcomes describe construction-time hardness metadata rather than retrieval-augmented baseline performance.}
\label{fig:dataset-profile}
\end{figure*}

LitTraceQA is released as a benchmark and shared-task development set over a
paper metadata pool. The public split provides the task contract: questions,
gold relevant papers, evidence annotations, and answers. PDFs are not
redistributed with the dataset; systems operate over the released metadata and
any paper access available under the original publishers' terms. To scale beyond
the compact public development split, we use a generate--ground--challenge
construction process. Paper metadata and local text caches provide source
material; an open-book generator proposes candidate questions and gold
annotations; automated grounding checks verify that evidence is attached to the
declared papers; and closed-book challenger models test whether the question can
be answered without access to the papers. Figure~\ref{fig:dataset-profile}
summarizes the scale and composition of the resulting collection; the paragraphs
below detail the individual construction stages.

\paragraph{Source selection.}
The metadata pool provides paper identifiers, titles, authors, abstracts,
venues, years, and available source links. The larger local source pool contains
27,487 paper metadata records from recent machine learning, computer vision, and
natural language processing venues. Candidate papers are selected from title and
abstract information, and a paper must have a local text cache before it can
support an internally generated instance. This requirement keeps generation tied
to inspectable paper content rather than to title-only metadata.

\paragraph{Open-book generation.}
For each candidate instance, the generator receives paper text and a target
evidence category. It produces a self-contained question, a set of gold papers,
typed evidence items, and a final answer. The target evidence category rotates
across tables, figures, text spans, equations or algorithms, and citation
contexts. Answer formats are constrained to multiple choice and structured
tables, which makes automatic checking more feasible than open-ended free-form
answers.

\paragraph{Grounding and filtering.}
After generation, each candidate passes evidence membership and grounding
checks. Evidence must refer to a declared gold paper, and distinctive text or
values should be recoverable from the local paper text. The same question is
then given to closed-book challenger models without paper context. Instances
whose answers can be guessed reliably without retrieval are filtered out; the
retained collection records whether all challenger models were wrong or whether
one challenger model answered correctly.

\begin{figure*}[t]
\centering
\includegraphics[width=\textwidth]{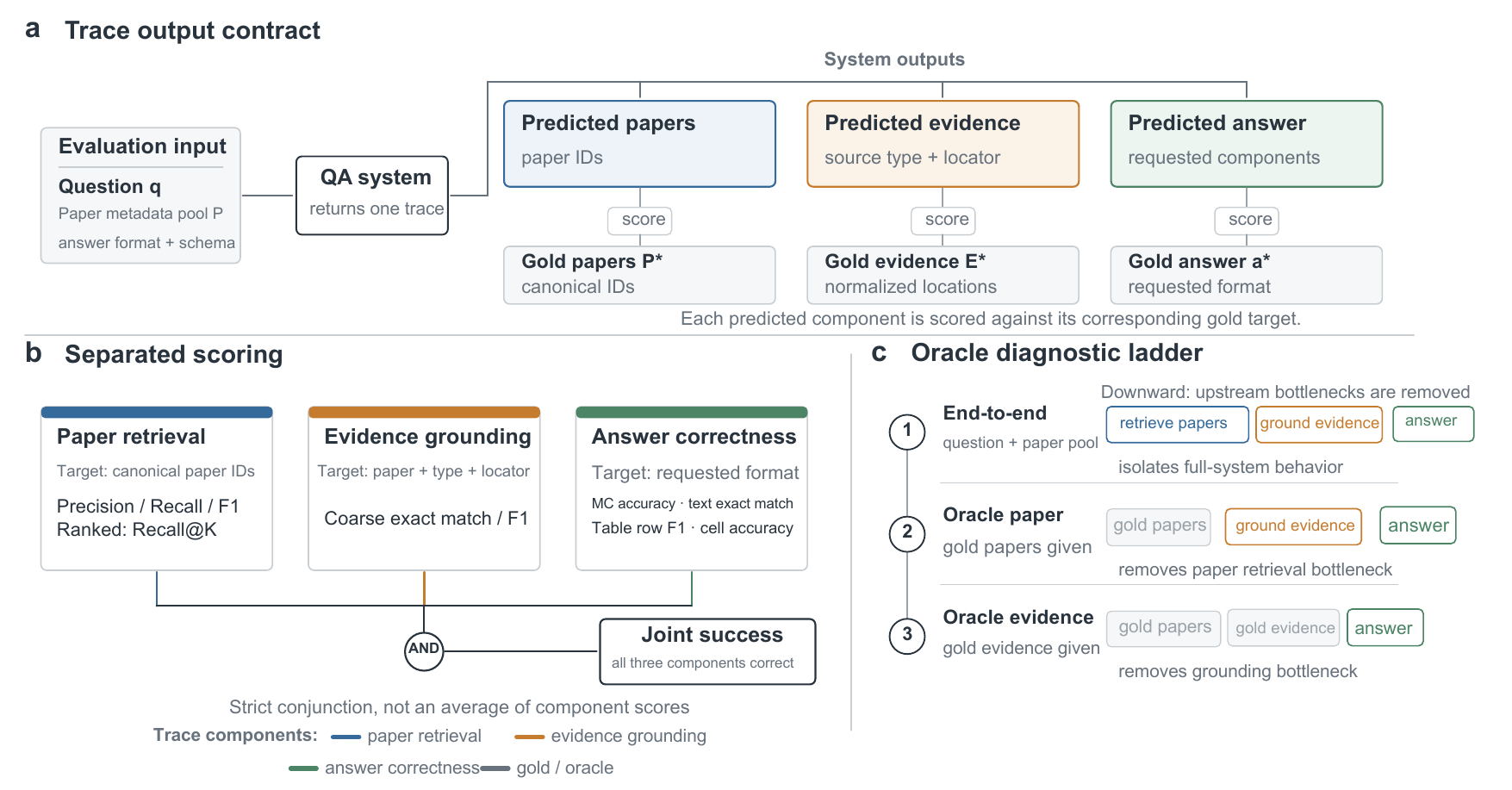}
\caption{LitTraceQA evaluation protocol. Systems return a trace consisting of paper identifiers, evidence locations, and answers. Metrics are reported separately for paper retrieval, evidence grounding, and answer correctness, while strict joint success requires all components to be correct. Oracle-paper and oracle-evidence settings provide a diagnostic ladder for isolating retrieval, grounding, and answer-generation failures.}
\label{fig:evaluation-protocol}
\end{figure*}

\paragraph{Annotation schema.}
Each record stores the question, primary evidence type, answer type, gold
papers, evidence items, answer, generation metadata, model attempts, difficulty
label, and verification metadata. Evidence items include a paper identifier, a
source type, evidence text or value, a locator, and grounding strings. The
locator field is flexible in the current collection because different evidence
types require different anchors, but release-quality evaluation will require
more normalization. Figure~\ref{fig:evaluation-protocol} shows how this trace
contract maps to component-level scores and oracle diagnostic settings.

\section{Dataset Characteristics}

We aggregate the local annotation files and collapse exact repeated question
texts from cumulative intermediate checkpoints before reporting corpus-level
statistics. The resulting final collection contains \MPQASamples{}
unique-question records over \MPQAUniquePapers{} unique gold papers and
\MPQAGoldLinks{} gold-paper links. Figure~\ref{fig:dataset-profile} gives the
compact table-and-chart summary of question scope, primary evidence, answer
format, and closed-book hardness. Figure~\ref{fig:venue-dist} separately reports
venue coverage, which is not represented in the record-level corpus profile.

\begin{figure}[t]
\centering
\small
\begin{tabular}{p{0.22\linewidth}rp{0.48\linewidth}}
\toprule
Venue & Gold-paper links (\%) & Relative frequency \\
\midrule
\MPQAVenueBars
\bottomrule
\end{tabular}
\caption{Venue distribution by gold-paper link. The current collection is concentrated in recent vision and NLP venues, indicating a clear target domain but also a venue-balance limitation.}
\label{fig:venue-dist}
\end{figure}

\paragraph{Coverage.}
The collection contains all five intended primary evidence types: 1,339
table-based questions (26.9\%), 1,132 figure-based questions (22.7\%), 933
text-span questions (18.7\%), 819 equation-or-algorithm questions (16.5\%),
and 755 citation-context questions (15.2\%). Evidence items are more
mixed than primary task labels: a table-centered question may still require
supporting prose, and a figure-centered question may require caption text or
citation context.

\paragraph{Answer formats and hardness.}
The final collection contains 3,029 multiple-choice instances
(60.8\%) and 1,949 structured-table instances (39.2\%). The difficulty labels
show that 3,080 instances (61.9\%) were answered incorrectly by all closed-book
challenger models, while 1,898 instances (38.1\%) were answered correctly by exactly one
challenger. These labels do not replace
retrieval-augmented evaluation, but they provide an initial hardness check:
many questions are not reliably answerable from model memory alone.

\paragraph{Source diversity.}
The collection uses \MPQAUniquePapers{} unique gold papers and
\MPQAGoldLinks{} gold-paper links. Most gold-paper links come from 2025 papers,
with a smaller number from 2024. The venue distribution is concentrated in ICCV,
NAACL, and ACL, with additional links from ICML, ECCV, EMNLP, NeurIPS, CVPR, and
ICLR (Figure~\ref{fig:venue-dist}). This concentration is a property of the
current annotation collection and should be treated as a limitation or as a
deliberate domain focus.

\section{Evaluation Protocol}

Figure~\ref{fig:evaluation-protocol} summarizes the output contract, separated
metrics, strict joint criterion, and oracle diagnostic settings. LitTraceQA
should be evaluated with separated metrics rather than a single answer score.
\textbf{Paper retrieval} is evaluated over canonical paper
identifier sets, with question-level precision, recall, and F1 macro-averaged
across questions; ranked variants such as Recall@K, MRR, MAP, or nDCG can be
reported when systems return ordered lists. \textbf{Evidence grounding} is
evaluated over normalized evidence locations comprising the paper identifier,
evidence source type, page number, and an object identifier when the evidence
type supports one. Coarse exact match checks complete locator tuples, while
set-level precision, recall, and F1 can supplement it when predictions contain
multiple evidence items. For table evidence, the object identifier is the table ID; for
figure evidence, it is the figure ID; for text spans, equation-or-algorithm
evidence, and citation contexts, the page-level location is used as the coarse
grounding target. \textbf{Answer correctness} follows the requested answer
components: accuracy for multiple-choice answers, exact match for short
free-form answers in the public evaluator, and row-level F1 plus macro- and
micro-averaged cell accuracy for structured tables. \textbf{Joint success}
requires the paper, evidence, and answer components all to be correct; it is a
strict conjunction rather than an average of component scores.

This separation is essential for diagnosis. A model that answers correctly but
retrieves the wrong paper has not solved retrieval-grounded QA. A model that
retrieves the right paper but cites an irrelevant evidence item has not solved
grounding. A model that retrieves and grounds correctly but produces the wrong
table value has failed answer generation. Reporting these stages separately
allows LitTraceQA to distinguish retrieval failures, grounding failures, and
reasoning or formatting failures.

The benchmark can support several evaluation settings. In the full end-to-end
setting, the model receives only the question and the paper pool. In an oracle
paper setting, the model receives the gold papers and must identify evidence and
answer. In an oracle evidence setting, the model receives the gold evidence and
must answer. These settings form a diagnostic ladder: end-to-end performance
measures complete capability, while oracle variants isolate the contribution of
retrieval and evidence localization. They define diagnostic evaluation
conditions rather than additional model results.

\section{Corpus Analysis}

The corpus statistics suggest three findings about the current state of the
benchmark. \paragraph{Finding 1: LitTraceQA mostly measures cross-paper
retrieval, but not exclusively.} \MPQAMultiPaper{} instances require at least
two gold papers, accounting for \MPQAMultiPaperPct{} of the final
collection. The remaining \MPQASinglePaper{} instances (\MPQASinglePaperPct{})
are still useful for paper-internal grounding, but they should be reported
separately or filtered when the evaluation claim is specifically about
multi-paper reasoning.

\paragraph{Finding 2: Evidence grounding is heterogeneous.}
The collection contains 5,673 text-span evidence items (43.5\%), 2,103 table
evidence items (16.1\%), 1,905 citation-context evidence items (14.6\%), 1,720
equation-or-algorithm evidence items (13.2\%), and 1,647 figure evidence items
(12.6\%). This distribution means the benchmark does not reduce scientific QA
to paragraph extraction. It requires models to handle multiple paper artifact
types and to connect them to answer formats.

\paragraph{Finding 3: Closed-book hardness filtering changes the benchmark
target.}
Every final record contains verification metadata. The hardness labels
show 3,080 instances (61.9\%) for which all challenger models were wrong
without paper context and 1,898 instances (38.1\%) for which one challenger was
correct. This supports the intended evaluation target: the benchmark should
reward retrieval-grounded systems rather than systems that answer from memory.
At the same time, these labels are not leaderboard results because the
challenger models were not evaluated with paper retrieval.

\paragraph{Finding 4: The release needs balancing and normalization.}
The final collection has no duplicate question text under the
current duplicate check, but it contains 918 repeated gold-paper sets,
\MPQASinglePaper{} single-paper instances (\MPQASinglePaperPct{}), and 72
instances with fewer than two evidence items (1.4\%). Venue coverage is also
uneven. These are not flaws
in the task definition, but they are release decisions: the authors should
either define filtered subsets or explicitly position the benchmark as covering
both single-paper grounding and multi-paper retrieval-grounded QA.

\section{Limitations and Release Requirements}

The current collection is strong enough to support a benchmark-development
paper, but several items remain necessary for a final benchmark release. First,
the flexible evidence locator schema should be normalized so that grounding can
be scored automatically across tables, figures, equations, algorithms, citation
contexts, and text spans. Second, the benchmark should define official splits or
public/hidden evaluation partitions. Third, the single-paper subset should be
handled deliberately, either as a separate diagnostic subset or as data to
filter for a strictly multi-paper track. Fourth, licensing and redistribution
status must be documented for metadata, paper text, and released annotations.
Fifth, baseline retrieval-augmented systems should be evaluated under the same
paper, evidence, answer, and joint metrics proposed above.

Human quality control is another open requirement. The current collection
contains automated generation, grounding, and challenger-verification metadata,
but a submission-ready benchmark should also document human review or expert
spot-check procedures. If multiple annotators are used, inter-annotator
agreement should be reported for evidence type, evidence locator, answer format,
and final answer correctness. If the benchmark remains primarily
semi-automated, the release should disclose which checks are automatic and which
properties are not guaranteed.

\section{Conclusion}

LitTraceQA reframes scientific-paper QA as a traceable retrieval-grounding
problem. Instead of asking only whether a system can produce an answer, it asks
whether the system can find the relevant papers, locate typed evidence, and
answer in the requested format. The public development split contains 55
examples with gold paper, evidence, and answer annotations. The larger local
annotation collection contains \MPQASamples{} final unique-question records over
\MPQAUniquePapers{} unique gold papers, covers five evidence categories, and
records closed-book hardness metadata for every instance. The next step is to
turn this annotation collection into an official benchmark release with
normalized locators, official splits, documented licensing, human
quality-control evidence, and baseline retrieval-augmented evaluations.

\bibliography{refs}
\bibliographystyle{iclr26/iclr2026_conference}

\appendix
\section{Reproducibility Notes}

The corpus statistics in this draft are generated from the local LitTraceQA
annotation directory. Before statistics are exported, exact repeated question
texts from cumulative intermediate checkpoints are collapsed. The main paper
therefore reports the final unique-question collection: records with different
question text are counted as distinct instances, while exact duplicate question
texts are removed from corpus-level counts. The generated artifacts are stored
under the local analysis directory as JSON, CSV, Markdown, and LaTeX macro
files.

\section{Representative Annotation Examples}
\label{app:annotation-examples}

This appendix gives compact but complete paper-facing annotation cards for the
examples summarized in Table~\ref{tab:task-examples}. For readability, we omit
local file names and use sample labels rather than internal collection
filenames. Each card reports the collected task fields, gold papers, evidence
items, answer object, generation metadata, verification metadata, and the
wrong-answer behavior observed during closed-book challenger validation. For
space, raw challenger outputs are rendered as compact answer strings: the
predicted values, choices, and cited works are preserved, while surrounding
JSON formatting is omitted.

\subsection{Sample A: Figure-Grounded Robot Manipulation Comparison}

\paragraph{Collected task fields.}
Task family: \texttt{multi\_paper}. Primary evidence type: \texttt{figure}.
Answer type: \texttt{table}. Difficulty after verification:
\texttt{all\_models\_wrong}. The question asks: ``For these two
diffusion-based robot manipulation policy papers, report the average
success-rate improvement each method claims and the exact number of real-world
evaluation tasks used to compute the real-world figure: FVP (built on DP3
across its real tasks) versus the Image-to-Sphere Policy across its real-world
tasks.''

\paragraph{Gold papers.}
\begin{itemize}[leftmargin=*]
    \item \texttt{iccv2025\_00015}, \emph{4D Visual Pre-training for Robot Learning}, ICCV 2025.
    \item \texttt{neurips2025\_00005}, \emph{3D Equivariant Visuomotor Policy Learning via Spherical Projection}, NeurIPS 2025.
\end{itemize}

\paragraph{Collected evidence.}
\begin{enumerate}[leftmargin=*]
    \item \texttt{ev\_001}: \texttt{iccv2025\_00015}, source type
    \texttt{figure}, page 1, Abstract/Figure~1 context. Evidence value:
    ``Across twelve real-world manipulation tasks, FVP boosts the average
    success rate of 3D Diffusion Policy (DP3) for these tasks by 28\%.''
    Search needles include ``boosts the av-'' and ``twelve real-world
    manipulation tasks''.
    \item \texttt{ev\_002}: \texttt{iccv2025\_00015}, source type
    \texttt{text\_span}, page 1, Abstract. Evidence value repeats the
    task-count and 28\% improvement span for FVP. Search needles include
    ``for these'' and ``tasks by 28\%''.
    \item \texttt{ev\_003}: \texttt{neurips2025\_00005}, source type
    \texttt{text\_span}, page 2, Introduction contributions. Evidence value:
    ``We validate our method through extensive experiments, achieving an
    average success rate improvement of 11.6\% over twelve simulation tasks
    and 42.5\% across four real-world tasks.'' Search needles include
    ``42.5\% across four real-world tasks''.
    \item \texttt{ev\_004}: \texttt{neurips2025\_00005}, source type
    \texttt{figure}, page 1, Figure~1 caption. Evidence value identifies the
    Image-to-Sphere Policy as an SO(3)-equivariant policy learning framework
    based on a single eye-in-hand RGB image. Search needles include
    ``first SO(3)-equivariant'' and ``eye-in-hand RGB image''.
\end{enumerate}

\paragraph{Gold answer object.}
The table schema has row key \texttt{method} and answer columns
\texttt{real\_world\_improvement} and \texttt{num\_real\_world\_tasks}. The
rows are: FVP (on DP3), 28\%, 12; and Image-to-Sphere Policy (ISP), 42.5\%, 4.

\paragraph{Generation and verification traces.}
The sample was generated by \texttt{us.anthropic.claude-opus-4-8} at
\texttt{2026-07-13T19:37:16.875054+00:00}. The verifier is
\texttt{generate\_loop}; verification at
\texttt{2026-07-13T19:37:16.875095+00:00} marked Claude, ChatGPT, and Gemini
as failures under the model mapping
\texttt{us.anthropic.claude-opus-4-8}, \texttt{openai.gpt-5.5}, and
\texttt{google.gemma-4-31b}. The challenger attempts each parsed only 1/5
checked table cells correctly.
\begin{itemize}[leftmargin=*]
    \item \textbf{Claude raw wrong output}: FVP about 25\% over 4 real-world
    tasks; Image-to-Sphere Policy about 18\% over 6 real-world tasks.
    \item \textbf{ChatGPT raw wrong output}: FVP about 11 percentage points
    over DP3 with 6 real-world tasks; Image-to-Sphere Policy about 17
    percentage points with 4 real-world tasks.
    \item \textbf{Gemini raw wrong output}: FVP approximately 15.3\% over 18
    real-world tasks; Image-to-Sphere Policy approximately 20\% over 10
    real-world tasks.
\end{itemize}
\emph{Error analysis.} All three models recognized that the question asks for
two robot policy papers, but they answered from plausible prior-like memory
rather than the cited figure/abstract evidence. The failure is mainly an exact
evidence-grounding error: the correct cells require copying 28\%/12 from the
FVP evidence and 42.5\%/4 from the Image-to-Sphere evidence.

\subsection{Sample B: Cross-Paper Retrieval Metric Alignment}

\paragraph{Collected task fields.}
Task family: \texttt{multi\_paper}. Primary evidence type: \texttt{table}.
Answer type: \texttt{table}. Difficulty after verification:
\texttt{all\_models\_wrong}. The question asks for three headline retrieval
gains: Diff2I2P registration-recall improvement on 7-Scenes, DAC average mAP
gain over prior arts across four open-set 3D object retrieval datasets, and
the average R@1 margin by which BLiM with CPN outperforms previous state of
the art on four text-video retrieval benchmarks.

\paragraph{Gold papers.}
\begin{itemize}[leftmargin=*]
    \item \texttt{iccv2025\_00559}, \emph{Diff2I2P: Differentiable Image-to-Point Cloud Registration with Diffusion Prior}, ICCV 2025.
    \item \texttt{iccv2025\_00540}, \emph{Describe, Adapt and Combine: Empowering CLIP Encoders for Open-set 3D Object Retrieval}, ICCV 2025.
    \item \texttt{iccv2025\_00253}, \emph{Bidirectional Likelihood Estimation with Multi-Modal Large Language Models for Text-Video Retrieval}, ICCV 2025.
\end{itemize}

\paragraph{Collected evidence.}
\begin{enumerate}[leftmargin=*]
    \item \texttt{ev\_001}: \texttt{iccv2025\_00559}, source type
    \texttt{text\_span}, Abstract. Evidence value: ``achieving over 7\%
    improvement in registration recall on the 7-Scenes benchmark.'' Search
    needle: ``over 7\% improvement in registration recall on the 7-Scenes
    benchmark''.
    \item \texttt{ev\_002}: \texttt{iccv2025\_00540}, source type
    \texttt{text\_span}, Abstract. Evidence value: ``DAC significantly
    surpasses prior arts by an average of +10.01\% mAP on four open-set 3DOR
    datasets.'' Search needle: ``by an average of +10.01\% mAP on four
    open-set 3DOR''.
    \item \texttt{ev\_003}: \texttt{iccv2025\_00253}, source type
    \texttt{text\_span}, Abstract. Evidence value: ``our BLiM equipped with
    CPN outperforms previous state-of-the-art models by 6.4 R@1 on average.''
    Search needle: ``outperforms previous state-of-the-art models by 6.4 R@1
    on average''.
    \item \texttt{ev\_004}: \texttt{iccv2025\_00540}, source type
    \texttt{text\_span}, Abstract. Evidence value states that DAC synergizes
    a CLIP model with a multi-modal large language model; it records method
    context for the DAC row.
\end{enumerate}

\paragraph{Gold answer object.}
The table schema uses \texttt{method} as the row key and stores
\texttt{reported\_gain} plus \texttt{benchmark\_or\_setting}. The rows are:
Diff2I2P, over 7\%, registration recall on 7-Scenes; DAC, +10.01\% mAP,
average over four open-set 3DOR datasets; and BLiM+CPN, 6.4 R@1, average over
four text-video retrieval benchmarks.

\paragraph{Generation and verification traces.}
The sample was generated by \texttt{us.anthropic.claude-opus-4-8} at
\texttt{2026-06-30T21:29:51.484207+00:00}. The verifier is
\texttt{generate\_loop}; verification at
\texttt{2026-06-30T21:29:51.484241+00:00} marked Claude, ChatGPT, and Gemini
as failures. Each challenger parsed only 1/6 checked cells correctly.
\begin{itemize}[leftmargin=*]
    \item \textbf{Claude raw wrong output}: Diff2I2P about 13.5\% registration
    recall improvement; DAC roughly 4--5\% mAP; BLiM+CPN about 2.6 R@1.
    \item \textbf{ChatGPT raw wrong output}: Diff2I2P about 12.2 percentage
    points; DAC about 5.5 mAP points; BLiM+CPN about 2.7 R@1 points.
    \item \textbf{Gemini raw wrong output}: Diff2I2P 1.2\%; DAC 3.5\% mAP;
    BLiM+CPN 2.1 R@1.
\end{itemize}
\emph{Error analysis.} The wrong outputs have the right surface form--three
retrieval papers and three metric-like numbers--but none preserves the
abstract-level claims. This is a cross-paper alignment failure: each answer row
must be tied to a different paper, and approximate retrieval-metric memory is
not enough to recover over 7\%, +10.01\% mAP, and 6.4 R@1.

\subsection{Sample C: Inference-Time Adaptation Gains}

\paragraph{Collected task fields.}
Task family: \texttt{multi\_paper}. Primary evidence type:
\texttt{text\_span}. Answer type: \texttt{table}. Difficulty after
verification: \texttt{all\_models\_wrong}. The question asks for the headline
quantitative gains reported by two inference-time adaptation papers: the
average AP3D$\vert$R40 improvement in the Car category that DUO reports on
KITTI, and the cross-domain few-shot segmentation gain that Informative
Structure Adaptation reports when applied to the SSP baseline.

\paragraph{Gold papers.}
\begin{itemize}[leftmargin=*]
    \item \texttt{iccv2025\_00067}, \emph{Adaptive Dual Uncertainty Optimization: Boosting Monocular 3D Object Detection under Test-Time Shifts}, ICCV 2025.
    \item \texttt{iccv2025\_00063}, \emph{Adapting In-Domain Few-Shot Segmentation to New Domains without Source Domain Retraining}, ICCV 2025.
\end{itemize}

\paragraph{Collected evidence.}
\begin{enumerate}[leftmargin=*]
    \item \texttt{ev\_001}: \texttt{iccv2025\_00067}, source type
    \texttt{text\_span}, Introduction. Evidence value: ``achieving
    state-of-the-art results with average improvements of +2.2 AP3D$\vert$R40
    in the Car category.'' Search needles include ``average improvements of
    +2.2 AP3D$\vert$R40 in the Car'' and ``13 corruption shift types''.
    \item \texttt{ev\_002}: \texttt{iccv2025\_00067}, source type
    \texttt{text\_span}, Introduction. Evidence value records that the method
    is evaluated on KITTI with 13 corruption shift types.
    \item \texttt{ev\_003}: \texttt{iccv2025\_00063}, source type
    \texttt{text\_span}, Figure~1 overview. Evidence value: ``SSP + ISA 63.0
    $\rightarrow$ 70.3.'' Search needles include ``SSP + ISA'' and
    ``Informative Structure Adaptation''.
    \item \texttt{ev\_004}: \texttt{iccv2025\_00063}, source type
    \texttt{text\_span}, Figure~1 overview. Evidence value: ``PANet + ISA
    55.1 $\rightarrow$ 61.8.'' This auxiliary row records the same figure's
    cross-domain context.
\end{enumerate}

\paragraph{Gold answer object.}
The table schema uses \texttt{method} as the row key and stores
\texttt{paper\_focus}, \texttt{dataset\_or\_baseline}, and
\texttt{reported\_gain}. The rows are: DUO, monocular 3D object detection
test-time adaptation, KITTI Car category, +2.2 AP3D$\vert$R40; and ISA on SSP,
cross-domain few-shot segmentation, SSP baseline, 63.0 $\rightarrow$ 70.3.

\paragraph{Generation and verification traces.}
The sample was generated by \texttt{us.anthropic.claude-opus-4-8} at
\texttt{2026-06-30T20:47:43.331910+00:00}. The verifier is
\texttt{generate\_loop}; verification at
\texttt{2026-06-30T20:47:43.331982+00:00} marked all three challenger models
as failures. Each attempt parsed 0/3 checked cells correctly.
\begin{itemize}[leftmargin=*]
    \item \textbf{Claude raw wrong output}: DUO approximately +5.27\%
    AP3D$\vert$R40 in the Car category; ISA about +6.6\% mIoU on SSP.
    \item \textbf{ChatGPT raw wrong output}: DUO +5.75 AP3D$\vert$R40 on
    KITTI Car; ISA on SSP +7.9 mIoU.
    \item \textbf{Gemini raw wrong output}: DUO +2.4\% AP3D$\vert$R40; ISA
    +3.8\% mIoU over SSP.
\end{itemize}
\emph{Error analysis.} The DUO value is not a generic percentage gain; the
gold span states an average improvement of +2.2 AP3D$\vert$R40. The ISA row is
also not simply a computed mIoU delta; the annotation asks for the reported
overview value 63.0 $\rightarrow$ 70.3. The models therefore failed both exact
span retrieval and answer-format alignment.

\subsection{Sample D: Equation and Algorithm Grounding Across Spatial Vision}

\paragraph{Collected task fields.}
Task family: \texttt{multi\_paper}. Primary evidence type:
\texttt{equation\_algorithm}. Answer type: \texttt{table}. Difficulty after
verification: \texttt{all\_models\_wrong}. The question asks which exact
noise-feature-injection or noise-modeling mechanism each of three spatial
vision papers defines: the operator used to inject style features into 3DGS
latents, the forward-diffusion noising equation used by the text-to-video
backdoor model, and the noise-model assumption that the polarization burst
super-resolution work rejects.

\paragraph{Gold papers.}
\begin{itemize}[leftmargin=*]
    \item \texttt{iccv2025\_00043}, \emph{A3GS: Arbitrary Artistic Style into Arbitrary 3D Gaussian Splatting}, ICCV 2025.
    \item \texttt{iccv2025\_00212}, \emph{BadVideo: Stealthy Backdoor Attack against Text-to-Video Generation}, ICCV 2025.
    \item \texttt{iccv2025\_00223}, \emph{Benchmarking Burst Super-Resolution for Polarization Images: Noise Dataset and Analysis}, ICCV 2025.
\end{itemize}

\paragraph{Collected evidence.}
\begin{enumerate}[leftmargin=*]
    \item \texttt{ev\_001}: \texttt{iccv2025\_00043}, source type
    \texttt{equation\_algorithm}, Introduction/Method. Evidence value:
    ``we utilize Adaptive Instance Normalization (AdaIN) to inject features
    from the target style image into the latents of the 3D Gaussian scene.''
    Search needles include ``Adaptive Instance Normalization (AdaIN) to inject
    features'' and ``into the latents of the 3D Gaussian scene''.
    \item \texttt{ev\_002}: \texttt{iccv2025\_00212}, source type
    \texttt{equation\_algorithm}, Section 3.1 preliminaries. Evidence value:
    $z_t=\sqrt{\bar{\alpha}_t}z_0+\sqrt{1-\bar{\alpha}_t}\epsilon$,
    $\epsilon\sim\mathcal{N}(0,1)$. Search needles target the
    $\sqrt{\bar{\alpha}_t}z_0$ and
    $\sqrt{1-\bar{\alpha}_t}\epsilon$ terms.
    \item \texttt{ev\_003}: \texttt{iccv2025\_00223}, source type
    \texttt{equation\_algorithm}, Introduction. Evidence value: ``assumptions
    that noise follows an additive white Gaussian distribution in the Stokes
    vector components.'' Search needles include ``additive white Gaussian
    distribution'' and ``in the Stokes vector components''.
    \item \texttt{ev\_004}: \texttt{iccv2025\_00223}, source type
    \texttt{equation\_algorithm}, Introduction. Evidence value: ``based on a
    shot and read noise model of polarization sensors.'' Search needles include
    ``shot and read noise model of polarization sensors'' and ``with minimal
    assumptions''.
\end{enumerate}

\paragraph{Gold answer object.}
The table schema uses \texttt{paper} as the row key and stores
\texttt{noise\_or\_injection\_mechanism}. The rows are: A3GS uses AdaIN to
inject features into 3D Gaussian scene latents; BadVideo uses
$z_t=\sqrt{\bar{\alpha}_t}z_0+\sqrt{1-\bar{\alpha}_t}\epsilon$ with
$\epsilon\sim\mathcal{N}(0,1)$; and polarization burst SR rejects additive
white Gaussian noise in Stokes vector components and instead uses a shot/read
noise model of polarization sensors.

\paragraph{Generation and verification traces.}
The sample was generated by \texttt{us.anthropic.claude-opus-4-8} at
\texttt{2026-06-30T21:02:36.372763+00:00}. The verifier is
\texttt{generate\_loop}; verification at
\texttt{2026-06-30T21:02:36.372871+00:00} marked Claude, ChatGPT, and Gemini
as failures. Each parsed 0/3 checked cells correctly.
\begin{itemize}[leftmargin=*]
    \item \textbf{Claude raw wrong output}: AdaIN for style statistics; a
    standard DDPM-style equation using $x_t$ and $x_0$; a broad polarization
    noise description rather than the exact rejected assumption and replacement
    sensor model.
    \item \textbf{ChatGPT raw wrong output}: an expanded AdaIN formula
    $\mathrm{AdaIN}(c,s)$; $q(x_t\mid x_0)=\mathcal{N}(x_t;\sqrt{\bar{\alpha}_t}x_0,
    (1-\bar{\alpha}_t)I)$; a generic noise-model statement.
    \item \textbf{Gemini raw wrong output}: ``additive bias or AdaIN'' for the
    3DGS row; a standard forward-diffusion equation in $x_t$; a broad claim
    that the polarization paper rejects Gaussian/noisy assumptions.
\end{itemize}
\emph{Error analysis.} This sample exposes a partial-credit trap. The raw
responses mention concepts related to the gold answer, but the benchmark checks
the exact paper-row correspondence: A3GS requires AdaIN into 3DGS latents,
BadVideo requires the paper's $z_t$ noising equation, and the polarization
paper requires the contrast between additive white Gaussian noise in Stokes
components and a shot/read sensor-noise model. Generic diffusion notation or
unanchored noise descriptions do not satisfy the annotation.

\subsection{Sample E: Citation-Context Reasoning in LLM Compression}

\paragraph{Collected task fields.}
Task family: \texttt{multi\_paper}. Primary evidence type:
\texttt{citation\_context}. Answer type: \texttt{multiple\_choice}.
Difficulty after verification: \texttt{all\_models\_wrong}. The question asks
which work is cited by the self-calibration quantization paper for adapting
the Optimal Brain Surgeon framework to quantization, and which work is cited by
the inference-acceleration-bias paper for showing that knowledge distillation
causes a monotonic reduction in toxicity in GPT-2.

\paragraph{Gold papers.}
\begin{itemize}[leftmargin=*]
    \item \texttt{naacl2025\_00963}, \emph{Self-calibration for Language Model Quantization and Pruning}, NAACL 2025.
    \item \texttt{naacl2025\_01064}, \emph{The Impact of Inference Acceleration on Bias of LLMs}, NAACL 2025.
\end{itemize}

\paragraph{Collected evidence.}
\begin{enumerate}[leftmargin=*]
    \item \texttt{ev\_001}: \texttt{naacl2025\_00963}, source type
    \texttt{citation\_context}, Section 2.1 Model Compression--Quantization.
    Evidence value: ``In a separate line of work, Frantar and Alistarh (2022)
    adapt the OBS framework to quantization. GPTQ (Frantar et al., 2023)
    builds upon this work to enable second-order low-bit quantization for
    LLMs.'' Search needles include ``Frantar and Alistarh (2022) adapt the OBS
    framework to quantization''.
    \item \texttt{ev\_002}: \texttt{naacl2025\_01064}, source type
    \texttt{citation\_context}, Section 2 Related Work. Evidence value: ``Xu
    and Hu (2022) find that knowledge distillation causes a monotonic
    reduction in toxicity in GPT-2, though it shows only small improvements in
    reducing bias on counterfactual embedding-based datasets.'' Search needles
    include ``monotonic reduction in toxicity in GPT-2''.
    \item \texttt{ev\_003}: \texttt{naacl2025\_00963}, source type
    \texttt{citation\_context}, Section 2.1 Model Compression--Pruning.
    Evidence value: ``SparseGPT (Frantar and Alistarh, 2023) presents an
    approximate weight reconstruction approach, enabling efficient LLM pruning
    without compromising performance.'' This auxiliary citation context helps
    distinguish pruning from the OBS-to-quantization citation.
\end{enumerate}

\paragraph{Gold answer object.}
The answer options are: A, Frantar et al. (2023) / Gonçalves and Strubell
(2023); B, Dettmers et al. (2022) / Hong et al. (2024); C, Frantar and
Alistarh (2022) / Xu and Hu (2022); and D, LeCun et al. (1989) / Jaiswal et
al. (2024). The gold answer is C.

\paragraph{Generation and verification traces.}
The sample was generated by \texttt{us.anthropic.claude-opus-4-8} at
\texttt{2026-06-29T00:58:07.311535+00:00}. The verifier is
\texttt{generate\_loop}; verification at
\texttt{2026-06-29T00:58:07.311561+00:00} marked Claude, ChatGPT, and Gemini
as failures. All three parsed answer A.
\begin{itemize}[leftmargin=*]
    \item \textbf{Claude raw wrong output}: answer A, citing GPTQ/Frantar et
    al. (2023) for OBS-related quantization and Gonçalves and Strubell (2023)
    for the toxicity/distillation claim.
    \item \textbf{ChatGPT raw wrong output}: answer A, with cited-paper IDs
    resembling \texttt{frantar-etal-2023-gptq} and
    \texttt{goncalves-strubell-2023-compression}.
    \item \textbf{Gemini raw wrong output}: answer A, citing a Frantar 2023
    SparseGPT-style work and a Gonçalves 2023 toxicity/compression work.
\end{itemize}
\emph{Error analysis.} The models identified the broader literature areas but
missed the local citation relation. The self-calibration paper cites Frantar
and Alistarh (2022) as the work adapting OBS to quantization, while GPTQ is
described as building on that work; the bias paper cites Xu and Hu (2022) for
monotonic toxicity reduction. The error is therefore citation-attribution
drift, not a failure to understand model compression in general.

\end{document}